\documentclass[10pt, conference, compsocconf]{IEEEtran}
\usepackage{graphicx}
\usepackage{subfigure}
\usepackage{multirow}
\usepackage{makecell}
\usepackage{diagbox}
\ifCLASSINFOpdf
\else
\fi
\begin{document}
%
\title{Towards Document Image Quality Assessment: A Text Line Based Framework and A Synthetic Text Line Image Dataset}


\author{
\IEEEauthorblockA{Hongyu Li\\
CV Lab, AI Institute\\ Tongdun Technology\\
Shanghai, China\\
hongyu.li@tongdun.net}
\and
\IEEEauthorblockA{Fan Zhu\\
AI Lab\\
ZhongAn Technology\\
Shanghai, China\\
zhufan@zhongan.io}
\and
\IEEEauthorblockA{Junhua Qiu\\
CV Lab, AI Institute\\ Tongdun Technology\\
Shanghai, China\\
junhua.qiu@tongdun.net}
}


%


\maketitle

\begin{abstract}
	Since the low quality of document images will greatly undermine the chances of success in automatic text recognition and analysis, it is necessary to assess the quality of document images uploaded in online business process, so as to reject those images of low quality. In this paper, we attempt to achieve document image quality assessment and our contributions are twofold. Firstly, since document image quality assessment is more interested in text, we propose a text line based framework to estimate document image quality, which is composed of three stages: text line detection, text line quality prediction, and overall quality assessment. Text line detection aims to find potential text lines with a detector. In the text line quality prediction stage, the quality score is computed for each text line with a CNN-based prediction model. The overall quality of document images is finally assessed with the ensemble of all text line quality. Secondly, to train the prediction model, a large-scale dataset, comprising 52,094 text line images, is synthesized with diverse attributes. For each text line image, a quality label is computed with a piece-wise function. To demonstrate the effectiveness of the proposed framework, comprehensive experiments are evaluated on two popular document image quality assessment benchmarks. Our framework significantly outperforms the state-of-the-art methods by large margins on the large and complicated dataset.

\end{abstract}


%
\IEEEpeerreviewmaketitle

\begin{figure*}[!tb]
	\centering
	\includegraphics [width=1\textwidth]{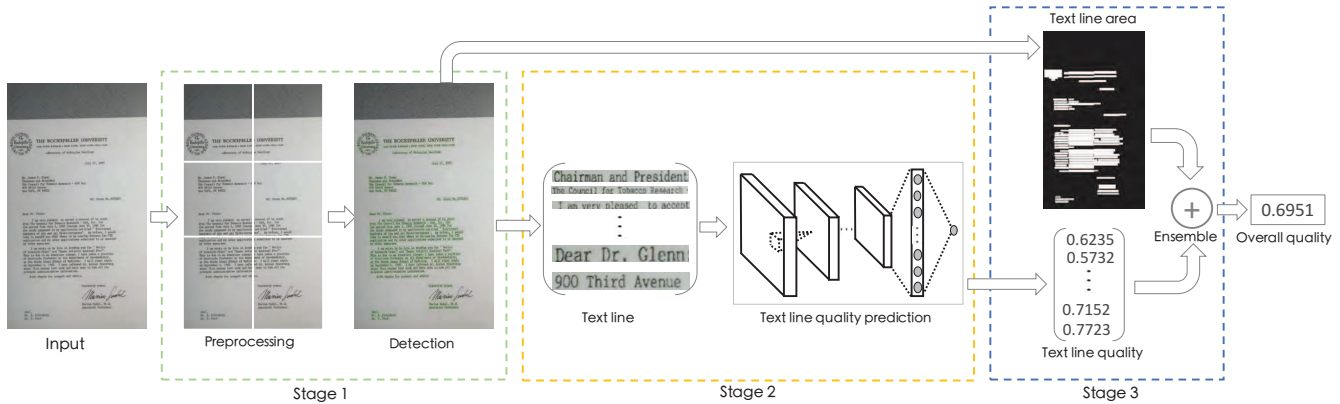}
	\caption{The proposed framework for document image quality assessment, consisting of three stages. Stage 1: text line detection, Stage 2: text line quality prediction, Stage 3: overall quality assessment.} 	
	\label{fig:pipeline}
\end{figure*}

\section{Introduction}
\label{sec:intro}
With the pervasive use of smart devices in our daily life, mobile captured document images are often required to be submitted in business processes of Internet companies, resulting that the amount of document images is rapidly increasing. Therefore, intelligent document recognition is becoming more significant for business process automation.
However, the performance of such recognition methods is greatly sensitive to document image quality.
The recognition accuracy of captured document images is often decreased with the low document image quality due to artifacts introduced during image
acquisition \cite{Ye2013Document}, which probably hinders the following business process severely.
For example, during online insurance claims, if a document image of low quality, submitted for claims, is not detected as soon as possible to require an immediate recapture,
critical information may be lost in business processes once the document is unavailable later.
Since the quality of document images uploaded by users is uneven, it is necessary to assess the quality of such images beforehand so as to reject those of low quality.

In past years, many algorithms have been developed for document image quality assessment (DIQA).
According to the difference of feature extraction, they can be categorized as two groups: metric-based
assessment and learning-based assessment.

The metric-based methods in \cite{Kumar2012Sharpness,Nayef2015Metric,Li2017CGDIQA} are usually based on hand-crafted features that have shown to correlate with the OCR accuracy. A non-parametric model, $\Delta DOM$, is adopted in \cite{Kumar2012Sharpness} to estimate the sharpness/blurriness of document images. Around 30 degradation-specific quality metrics have been proposed to measure noise and character shape preservation \cite{Nayef2015Metric}. In \cite{Li2017CGDIQA}, a novel feature, character gradient, is designed to describe document image quality. But such methods often pay more attention to one specific feature,  and consequently may perform poorly on complicated and heterogeneous document images.

The learning-based methods take advantage of learning techniques, such as \cite{Kang2014A,Peng2016Document}, to extract discriminant features for different types of document degradations. In \cite{Kang2014A},
the authors propose a deep learning approach for document
image quality assessment, which crops an image into patches and then uses the
CNN to estimate quality scores for selected patches.
However, the training procedure in these methods requires massive labelled samples that are scarce or unavailable.

Different from scene images, document images are naturally more concerned about text. Therefore, the document image quality can be considered as an ensemble of the degradation degree of all text areas in a document image.
Motivated by this consideration,
we propose a text line based framework for DIQA in this work, which is divided into three stages:  text line detection, text line quality prediction, and overall quality assessment.

The text line detection stage manages to find potential text lines as significant text areas. In this stage, any text line detector can be directly borrowed as long as it has good performance in recall and precision. Since the image that CNN-based detectors can accept as input is generally fixed-size, it is required to resize document images before detection. As a consequence, it may result in poor bounding boxes for small text lines.
To solve this issue, we divide a large document image into multiple small segments before detection.

In the second stage, the quality score is predicted respectively for each detected text line with a pretrained prediction model.
The prediction model is composed of the backbone layers, which can be transplanted from popular CNNs, and an auxiliary regression layer that is added on the top of the backbone. Our strategy is to train this model on text line images, with an advantage of being insensitive to background clutter and noise.

In the final stage, the overall quality of document images is assessed in an ensemble way. The ensemble strategy is realized through a mapping function from the text line quality to the overall document quality.
It is observed that document image quality is as well related to the size of text lines. As a result, the text line area is supposed to be taken into account in the mapping function.

In addition, to train the text line quality prediction model, it is necessary to collect enough text line samples containing text quality labels. Unfortunately, such a dataset does not exist for now. The publicly available datasets in \cite{Kumar2012Sharpness,Nayef2015SmartDoc} are very small, unsuitable for training deep neural networks, on one hand. On the other hand, they have only ground truth quality for documents, not for text lines. In this work, to fill the aforementioned research
gaps to some extent, we synthesize a large number of text lines that contain both Chinese and English characters with diverse attributes. To simulate real document images, each text line image is blurred through a Gaussian filter with a random standard deviation and rotated by a random small degree. The key and difficulty in this data synthesis is how to label text line samples with the ground truth quality. To do this, we design a piece-wise function with respect to the Gaussian standard deviation to model the ground truth quality.

Our main contributions can be summarized as follows:

\begin{enumerate}
	\item We propose a text line based framework for DIQA,
	and conduct extensive experiments to show that the framework
	provides significantly better results in assessing document image quality on the large and complicated dataset. We
	believe this framework is general and applicable to a variety of heterogeneous document images.
	\item To facilitate the study of DIQA,
	we establish a large-scale dataset comprising 52,094 text line images with quality labels and ground truth texts, where the generated quality labels are basically in line with human perception.
	This dataset can serve as a
	benchmark and be employed to train and validate new algorithms for either
	text line quality prediction or segmentation-free optical character recognition.
\end{enumerate}

\section{A Text Line Based Framework}
\label{sec:Framework}

The most striking characteristic which differentiates document
images from scene images is text. As a consequence, document image quality assessment can
be assumed as measuring the degradation degree of text \cite{Li2017CGDIQA}. Based on this assumption, we propose a text line based framework for document image quality assessment.

A high-level overview of our framework is illustrated in Fig. \ref{fig:pipeline}. In the framework, an image is first fed into a text line detector to find significant and valid text lines and then text line quality is predicted with a CNN-based prediction network. The overall document image quality is eventually assessed with an ensemble strategy through bringing all text line quality together as a group.
This framework can be divided into three stages: text line detection, text line quality prediction, and overall quality assessment.
Different stages of the proposed framework are described in detail in the consecutive subsections.

\subsection{Text Line Detection}

Text line detection is the first stage in the proposed framework, in which we aim to find meaningful text lines from the input document image. Here, a text line is defined as a text area that contains significant attributes: characters or words. Since a document image can break down into its constituents - text lines, it is expected to replace the document image with text lines during document image quality assessment. To well extract text lines from the document image, a stable and accurate detector must be selected with care, given that ambiguous text often appears  in document images.

Comprehensive reviews about text line detection can be found
in the survey papers \cite{Ye2015Text}.
Previous text line detection approaches \cite{Tian2016Detecting,Buta2015FASText} have already obtained promising performances on various benchmarks and deep neural network based algorithms \cite{Tian2016Detecting,LiaoSBWL17} are becoming the mainstream in this field.
In the work \cite{Tian2016Detecting}, the connectionist text proposal network (CTPN) is proposed to
accurately localize text lines in scene images.
By exploring rich context information of image, CTPN is powerful in detecting extremely
ambiguous text and works reliably on multi-scale and multi-language
text without further post-processing.
Due to its high efficiency and good performance in text line detection, CTPN is selected as the text line detector and is directly utilized without finetuning in this work.

\begin{figure}[b]
	\centering
	\includegraphics [width=0.45\textwidth]{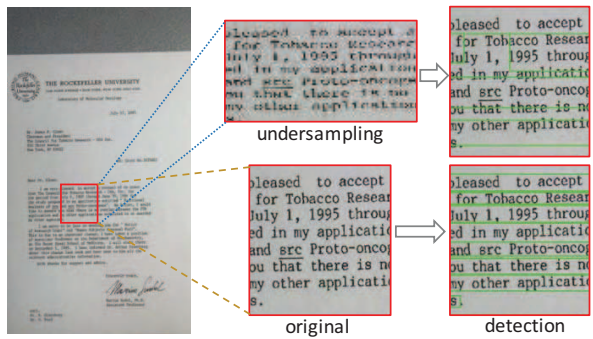}
	\caption{Detection with or without undersampling. Detected bounding boxes deteriorate after undersampling.} 	
	\label{fig:crop}
\end{figure}

It is worth noting that the size of a document image is generally much larger than the image size accepted by the deep convolutional neural networks. Moreover, the accepted size is required to be fixed in CNNs. To meet this requirement, the document image usually needs to be resized in the undersampling way before detection, resulting that small text may become blurry and even illegible after undersampling. That will eventually lead to the inaccurate localization of text lines, against the subsequent quality prediction. For example, once the document image in the left of Fig. \ref{fig:crop} is reduced with undersampling, text in the red box is becoming very small, around 7 pixels high, which severely deteriorates the performance of text line detection. However, it is fortunate that detection without undersampling is still reliable and desirable, as demonstrated in the bottom row of Fig.\ref{fig:crop}.

Consequently, to avoid such detection deterioration, we will employ a divide-and-conquer strategy in order to take advantage of small image divisions in place of the large input document image during text line detection. As illustrated in the preprocessing step of Fig.\ref{fig:pipeline}, the document image is first divided into several small segments with a reasonable size before detection, which can prevent harmful undersampling happening. Specifically, in the following experiments, each document image was divided into 4 equal pieces along x-axis and 6 equal pieces along y-axis, and 24 segments was totally collected for detection. Although a long and whole text line may be broken to pieces in the dividing process, it will not change the overall quality of text lines \textit{since the semantic content is entirely unrelated to the text line quality and is insignificant during the quality prediction.}
In fact, the detection deterioration will not usually happen to detectors based on hand-crafted features, where the dividing operation is unnecessary.

In addition, in the proposed framework, it is easy to replace CTPN with any other detectors in Stage 1, as long as the substitute has better efficacy and efficiency no matter whether it makes use of hand-crafted or deep features. It is worth emphasizing that document image quality assessment can benefit from text line detection effectively inhibiting background clutter and noise in document images.

\subsection{Text Line Quality Prediction}
The second stage of our framework aims to predict the quality of text lines found in the previous stage. Here we straightforwardly cast text line quality prediction as a regression problem and output a predicted quality score for each text line.

It has been proved in \cite{Kang2014A} that deep features are effective in document image quality assessment, which inspires us to employ a deep neural network to conduct text line quality prediction.
In the quality prediction network, the early layers can be based on any standard architecture truncated before the classification layer. An auxiliary regression layer, whose output is a neuron, is added behind the early layers for estimation. In our method, the ResNet \cite{He:resnet:2016} is adopted as a base due to its excellent ability of feature representation, but other networks should also produce good results.

\subsubsection{Loss Function}

To predict text line quality, the estimation loss adopts Euclidean loss for quality regression to describe the difference between the predicted and ground truth quality. Specifically, the estimation loss $\mathcal{L}_{q} $ is defined as
\begin{eqnarray}
\mathcal{L}_{q} = \|\mathbf{Q} - \overline{\mathbf{Q}} \|_2^2,
\end{eqnarray}
where $\mathbf{Q}$ and $\overline{\mathbf{Q}}$  are respectively the predicted and ground truth quality.

\subsubsection{Training Strategy}
The initial weights are assigned to the convolutional
layers in the prediction network with a robust initialization method \cite{He:2015:DDR}. The regression layer is randomly initialized under a uniform
distribution in the range (-0.1, 0.1).
In order to optimize the prediction network, the SGD optimizer is utilized with a learning rate 5$e$-3. A weight decay of 1$e$-4 is applied to all layers.

The prediction network is trained on our synthetic text line image dataset that is introduced in detail in the following section.
Since this dataset is established with attribute diversity, the prediction model is insensitive to heterogeneous text line images and is thus applicable for heterogeneous document images.

\begin{figure}[tb]
	\centering
	\includegraphics [width=.25\textwidth]{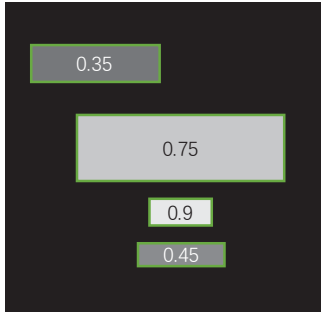}
	\caption{Overall quality assessment with the ensemble strategy. The text line quality is shown inside each text line, and the corresponding weights are 0.24, 0.62, 0.06, and 0.08. The overall quality obtained is 0.639.} 	
	\label{fig:overall}
\end{figure}

\subsection{Overall Quality Assessment}

In the last stage, the overall quality of a document image is assessed as the ensemble of quality of all text lines detected on it. It is observed that visual estimation involving document image quality is greatly affected by large text areas.
As a result, large text lines are supposed to contribute more to the overall document quality in our ensemble strategy.

The ensemble process is achieved through the weighted pooling of all text line quality on the basis of area ratio. In specific, we formulate the overall quality as follows,
\begin{equation}\label{eq:qual:img}
\hat{q}=\sum_jw_jq(j),
\end{equation}
where $w_j$ is a weight on the $j$-th text line of the image and is linearly proportional to the text line area. The weight is actually an area ratio: $w_j=\frac{R(j)}{\sum_k R(k)},$
where $R(j)$ and $\sum_k R(k)$ respectively represent the area of the $j$-th text line and the total area of all text lines in a document image.

Fig.\ref{fig:overall} presents an example about the overall quality assessment, where four text lines are detected and predicted respectively with quality of 0.35, 0.75, 0.9, and 0.45, from top to bottom. The corresponding weights calculated in terms of area ratio are 0.24, 0.62, 0.06, and 0.08. In this case, the overall quality $\hat{q}$ is \textit{0.639} according to Eq.\ref{eq:qual:img}, greater than the
median (0.6) of the text line quality. This proves that the ensemble process does highlight large text lines in overall quality assessment.

Note that although our overall assessment currently does not take into account other information, e.g. document layout or text line height, it is easy to incorporate such information into the proposed framework if necessary.

\section{A Synthetic Text Line Image Dataset}
\label{sec:data}

To train the text line quality prediction model, it is required to collect a large-scale set of text line images with quality labels. However, the public datasets, such as DocImg-QA dataset \cite{Kumar2012Sharpness}, are usually small and only provide labels for document images. Until now, there are no available datasets involving text line image quality. As a result, we need to establish such a dataset for training the prediction model, and will make it publicly available. Next, we will respectively describe the processes of text line image synthesis and quality label generation.

\subsection{Text Line Image Synthesis}

Here we introduce a novel way of synthesizing text line images. The procedure is briefly described in \textbf{Algorithm 1}.
\begin{figure}[!hb]
	\centering
	\includegraphics [width=.5\textwidth]{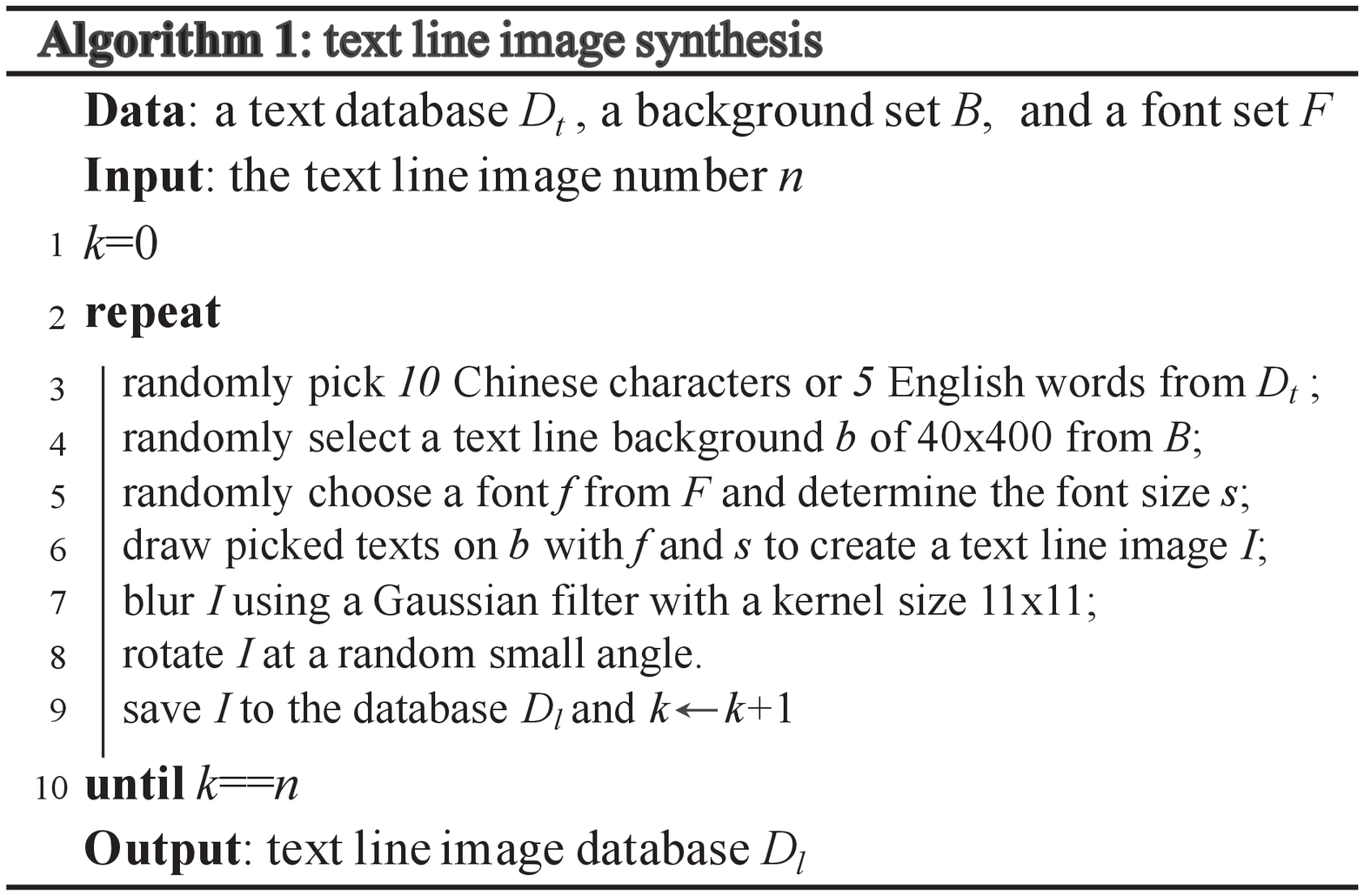}
	\label{alg:1}
\end{figure}

\begin{figure}[!hb]
	\centering
	\includegraphics [width=.4\textwidth]{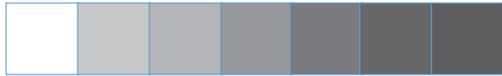}
	\caption{7 backgrounds with different grayscale.} 	
	\label{fig:background}
\end{figure}

For the purpose of synthesis, a text database needs to be created at first, comprising 8058 Chinese characters and 245 English words.
In each produced text line, there are about 10 Chinese characters or 5 English words. In effect, the semantic content involving text is insignificant in quality prediction, that is why we can randomly arrange text into text lines. To ensure data diversity, 6 popular fonts are applied during
synthesis.
The font size changes randomly from 28 to 35 for Chinese characters, and 35 to 45 for English words.
Because the background of real text lines consists of nearly solid color, we select 7 different grayscale values as text line backgrounds, as shown in Fig.\ref{fig:background}.

\begin{table}[b]
	\centering
	\caption{Attribute diversity for text line image synthesis.} \label{tab:diversity}	
	\begin{tabular}{l|c|c }
		\Xhline{1.2pt}	
		\textbf{Attribute}&\multicolumn{2}{c}{\textbf{Value}} \\
		\hline
		Font&\multicolumn{2}{c}{6 types}\\ \hline
		Background&\multicolumn{2}{c}{7 levels}\\ \hline
		Image size&\multicolumn{2}{c}{400*40} \\ \hline
		Kernel size&\multicolumn{2}{c}{11*11} \\ \hline
		Standard deviation&\multicolumn{2}{c}{0.5$\scriptsize{\sim}$4.5} \\ \hline
		Text&Chinese& English\\
		\hline
		Font size&28$\scriptsize{\sim}$35&35$\scriptsize{\sim}$45 \\ \hline
		Rotation angle&-2$\scriptsize{\sim}$2&-1$\scriptsize{\sim}$1 \\
		\hline
	\end{tabular}
\end{table}

In this work, our main concern is the degradation degree of text in images. According to the point of view in \cite{Ye2013Document,Bong2014,Bong2015}, blur is the most common issue in mobile captured images, which suggests that the blur degradation is more attractive and useful in practical applications.
Therefore, our synthetic data is produced under different levels of blur degradation.
To embody the quality difference between synthetic images, a random standard deviation $\sigma\in [0.5,4.5]$ is set for a Gaussian function to blur each text line image.
Besides, to simulate the realistic case, each text line is slightly rotated at an arbitrary angle between -2 and 2 degree for Chinese text as well as -1 and 1 degree for English text.

\begin{figure}[t]
	\centering
	\includegraphics [width=.5\textwidth]{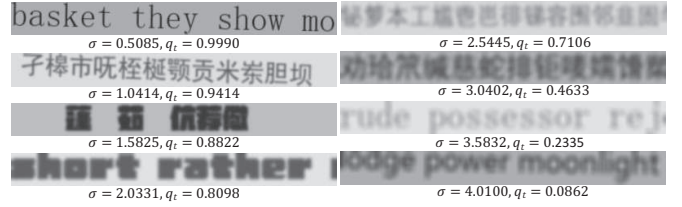}
	\caption{Examples of synthetic text lines with different attributes. Below each example are the quality label $q_t$ and the corresponding standard deviation $\sigma$.
	} 	
	\label{fig:syn-text lines}
\end{figure}

The attribute diversity is summarized in Table \ref{tab:diversity}. In total, 52,094 text line images are synthesized with various attributes, and some examples are shown in Fig.\ref{fig:syn-text lines}.
Note that there exist text lines deliberately not filled with sufficient characters/words in this dataset. This is done to make the prediction model more robust to text line detectors used in Stage 1. That is, even if the detected bounding boxes are not accurate enough, the text line quality can be still predicted well. Although the procedure of generating text line images of diverse characteristics is heuristic, it is the attribute diversity that makes our quality prediction model stronger in generalization, even for heterogenous text line images.

\subsection{Quality Label Generation}

In the synthetic data, blur is considered as the main factor of affecting text line image quality.
To generate quality labels for synthetic images, it is a must to build a link between the quality label $q_t$ and the degree of blur degradation controlled by a standard deviation $\sigma$.
Such relationship should be
a nonlinear mapping, $q_t=f(\sigma)$,
where the greater the standard deviation $\sigma$, the lower the image quality $q_t$.
To formulate this mapping, we design a piece-wise and continuous function of parameter $\sigma$,
\begin{equation}\label{eq-gt}
q_t=
\left\{
\begin{array}{lcl}
\frac{1}{1+(\sigma-0.5)*s_1} & & {0.5 \leq  \sigma < 1.5}\\
\frac{1}{1+p_1+(\sigma-1.5)*s_2} & & {1.5 \leq \sigma <2.5}\\
\frac{1}{1+p_2+(\sigma-2.5)*s_3} & & {2.5 \leq \sigma < 3.5}\\
\frac{1}{1+p_3+(\sigma-3.5)*s_4}  & & {3.5 \leq \sigma \leq 4.5}
\end{array} \right.
\end{equation}
where $p_i=\sum_{k=1}^i s_k$, $s_k$ is a scaling factor to adjust the effect of $\sigma$ on the quality label, and $q_t$ ranges from $\frac{1}{1+p_3+s_4}$ to 1. Here the quality is inverse proportional to $\sigma$ and decays faster when $\sigma$ approaches to 4.5. With this function, we can better model the quality variation of real-world text line, which is basically in line with human perception. By default, $s_1, \cdots, s_4$ are respectively set to be 0.115, 0.225, 1.515, and 17.145, with which the text line prediction model can perform better according to empirical studies.

\begin{figure}[!t]
	\centering
	\includegraphics [width=.45\textwidth]{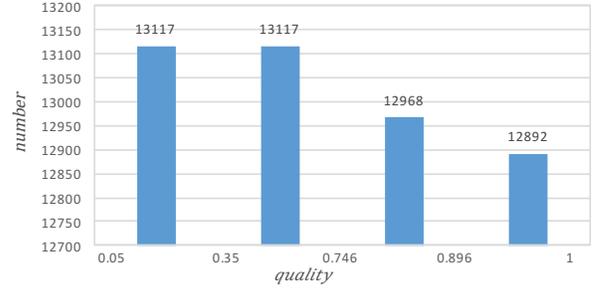}
	\caption{Distribution of synthetic data. There are 25,860 text line images labelled with quality larger than 0.746, and 13,117 images with quality less than 0.35.} 	
	\label{fig:stat}
\end{figure}

The standard deviation and generated quality label for each example in Fig.\ref{fig:syn-text lines} are presented below the image. With visual estimation, it is easy as well to conclude that the quality is becoming poor if the standard deviation is larger than 2.5.
The overall distribution of synthetic images along the ground truth quality we obtained is shown in Fig. \ref{fig:stat}, where there is the larger number 25,860 of high-quality images than 13,117 of low-quality images.

\section{Experiments}

In order to evaluate the proposed framework,
we conducted experimental analysis on two public benchmarks, \textbf{DocImg-QA} \cite{Kumar2013A} and \textbf{SmartDoc-QA} \cite{Nayef2015SmartDoc}, and compared it with state-of-the-art methods.

\subsection{Datasets and Protocols}

The DocImg-QA dataset contains a total
of 175 color images with resolution of 1840$\times$3264. To generate
different levels of blur degradations, 6-8 photos were taken with a smartphone for each of 25 documents involving machine-printed English characters.
In this dataset, three OCR engines: Finereader, Tesseract, and Omnipage, were run on each image to obtain the recognition accuracy as its ground truth quality.

SmartDoc-QA is a larger and more complicated dataset of 4,260
smartphone captured document images. 142 different images were
captured for one of 30 paper documents at different resolutions and distortions.
In SmartDoc-QA, there are Finereader and Tesseract recognition accuracies provided as the ground truth quality.

Like other works \cite{Li2017CGDIQA,Nayef2015SmartDoc}, we also use two metrics in this study for performance evaluation: the Linear Correlation Coefficient
(LCC) and the Spearman Rank Order Correlation Coefficient (SROCC).

Since different OCR engines have huge difference in accuracy, to avoid that the evaluation results are overwhelmingly dependent on a certain OCR engine, we claim to use the average recognition accuracy of different engines on either benchmark as the ground truth to compute LCC and SROCC.
In addition, to avoid the bias towards good results in terms of the document-wise evaluation protocol, we directly compute one LCC and one SROCC for all images in a dataset.

\subsection{Detector Selection}

Since the accurate localization of text lines can improve the performance of the proposed DIQA framework, it is significant and necessary to select an excellent detector in Stage 1. In this study, we tested three representative detectors: FASText\cite{Buta2015FASText}, TextBox\cite{LiaoSBWL17}, and CTPN\cite{Tian2016Detecting}, for comparative analysis.

For a fair comparison, in this test, all input images were first divided into 24 small segments, and each segment was separately handled with by the above-mentioned detectors to extract text lines. Besides, the quality prediction model and the overall ensemble strategy were kept same for each detector. The evaluation results with three detectors on two benchmarks are presented in Table \ref{tab:detector}. It is easy to conclude from the results that CTPN performs almost the best, except for a little lower at SROCC for DocImg-QA than FASText, and thus is more suitable for our framework.

\begin{table}[t]
	\centering
	\caption{Detector comparison on two benchmarks.} \label{tab:detector}	
	\begin{tabular}{l|c|c |c|c}
		\Xhline{1.2pt}	
		\multirow{2}{*}{Detector}&\multicolumn{2}{c|}{DocImg-QA(\%)}&\multicolumn{2}{c}{SmartDoc-QA(\%)} \\	
		\cline{2-5}
		&LCC& SROCC&LCC& SROCC\\
		\hline
		FASText&90.74 &\textbf{86.75} &69.02 &73.15 \\
		TextBox&89.18&83.73&63.55 &63.96 \\
		CTPN&\textbf{91.44}&85.67&\textbf{71.93 }&\textbf{74.64 }\\
		\hline
	\end{tabular}
\end{table}

\subsection{Dividing vs. Non-dividing}

As discussed previously, resizing a document image before detection will make the detected bounding boxes inaccurate and further deteriorate the performance of text line quality prediction. To solve this issue, our solution is to first divide a document image into several small segments and then pass each segment to the CNN-based detector. In this test, we contrasted the effect of non-dividing and dividing on the proposed framework, where all other parameters and operations were fixed.

Without the dividing operation, we directly reduced a document image to the resolution of 600$\times$900 acceptable to CTPN. In the dividing procedure, the document image was first partitioned into 24 small segments and each segment was then resized to the acceptable resolution. Experimental results on two benchmarks demonstrate that dividing is clearly superior to non-dividing in our framework, which corroborates our judgement that the resizing operation through undersampling damages the accurate localization of text lines and will further confuse text line quality prediction.

\begin{table}[ht]
	\centering
	\caption{Comparison between non-dividing and dividing.} \label{tab:comp:avg}	
	\begin{tabular}{l|c|c |c|c}
		\Xhline{1.2pt}	
		\multirow{2}{*}{Method}&\multicolumn{2}{c|}{DocImg-QA(\%)}&\multicolumn{2}{c}{SmartDoc-QA(\%)} \\	
		\cline{2-5}
		&LCC& SROCC&LCC& SROCC\\
		\hline
		Non-dividing&89.89 &82.60 &68.96 &72.70 \\
		Dividing&\textbf{91.44}&\textbf{85.67}&\textbf{71.93} &\textbf{74.64 }\\
		\hline
	\end{tabular}
\end{table}

\begin{table}[t]
	\centering
	\caption{6 groups of scaling factor configuration. $s_i$ and $Gi$ respectively represent the scaling factor and the group number. Each element corresponds to the value of a factor in a certain group.} \label{tab:scale}	
	\begin{tabular}{l|c|c|c|c|c|c}
		\Xhline{1.2pt}	
		&G1& G2&G3& G4&G5&G6\\
		\hline
		$s_1$&0.25 &0.115 &0.175 &0.325&0.325&0.25 \\
		$s_2$&0.5&0.225&0.365 &0.215&0.675&1.25 \\
		$s_3$&3.25&1.515&1.8 &2.46&1.335&7.5 \\
		$s_4$&15&17.145&16.65&16&16.665&90\\
		\hline
	\end{tabular}
\end{table}

\subsection{Scaling Factor for Label Generation}
\label{sec:exp:scale}
Since the scaling factor $s_i$ in quality label generation plays an important role in text line quality prediction, it is desired to find satisfactory scaling factors for our framework. To do this, we tested several reasonable configuration of scaling factors. There are three basic configuration principles: 1) set $s_1<s_2<1<s_3<s_4$, which can better reflect the fact that the quality of synthetic text line images decays more rapidly along with the increase of the standard deviation $\sigma$,
2) $s_1$ needs to be small enough so that text line images produced focus on the range of high quality when $\sigma<1.5$, 3) to guarantee that the computed low quality falls below a specified threshold (e.g., 0.3), $s_4$ should be large enough.

\begin{figure} [hb]
	\centering
	\includegraphics[width=0.48\textwidth]{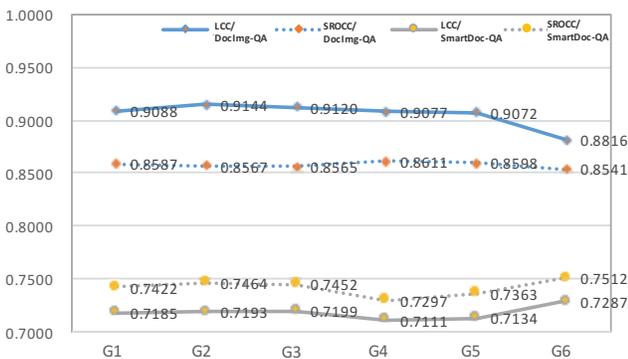}
	\caption{LCC and SROCC on two benchmarks under different scaling factor configuration listed in Table \ref{tab:scale}.} \label{fig:scale}
\end{figure}

Six groups $\{G1, \cdots, G6\}$ of scaling factor configuration are listed in Table \ref{tab:scale}. With each configuration, we respectively computed the LCC and SROCC on two benchmarks and presented them in Fig.\ref{fig:scale}. Experimental results show the second group $G2$ can achieve relatively better performance on both datasets, therefore we prefer to use it as default in practical applications.

\begin{table*}[!t]
	\centering
	\caption{Comparison with other DIQA methods on two benchmarks. The proposed framework with either weighted pooling or median significantly outperforms all the previous state-of-the-art methods on the large and complicated dataset, SmartDoc-QA.} \label{tab:comp:all}	
	\begin{tabular}{l|l|c|c|c|c|c}
		\Xhline{1.2pt}		
		Dataset&Protocol&MetricNR&Sharpness&CG-DIQA&Proposed+WP&Proposed+Median\\
		\hline
		\multirow{2}{*}{SmartDoc-QA}	&LCC(\%) &N/A&62.42&62.50&71.93&\textbf{74.33}\\
		&SROCC(\%) &N/A &59.64&63.05&74.64&\textbf{75.72}\\
		\hline
		\multirow{2}{*}{DocImg-QA} &LCC(\%) &88.67&N/A&90.63&\textbf{91.44}&90.89\\
		&SROCC(\%) &82.07&N/A&85.65&85.67&\textbf{87.23}\\
		\hline
	\end{tabular}
\end{table*}

\subsection{Overall Assessment Analysis}
In the ensemble strategy, the weighted pooling can be replaced with any other method. In this study, we also propose another simple and easy-to-implement way for overall quality assessment, i.e., picking the median among predicted quality scores of all text lines. It is obvious that the \textit{median} method is more effective for the document full of text lines with close quality. But the weighted pooling takes into full consideration the difference between text line scales during overall quality assessment, which accords more with visual estimation of document image quality.

\begin{figure}[h]
	\centering
	\subfigure[$\hat{q}_{wp}=0.3225$, $\hat{q}_{med}=0.9787$]{
		\label{fig:median1}
		\includegraphics [width=.45\textwidth]{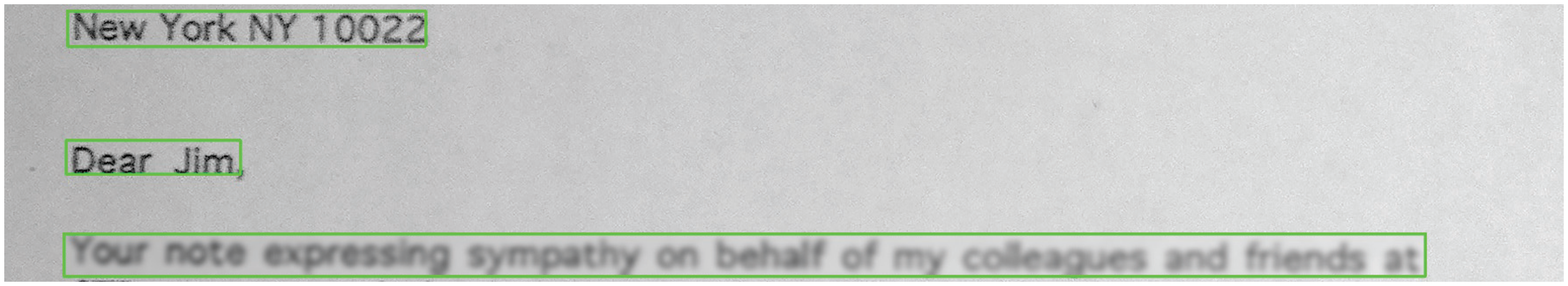}  }
	\subfigure[$\hat{q}_{wp}=0.3736$, $\hat{q}_{med}=0.1669$]{
		\label{fig:median2}
		\includegraphics [width=.45\textwidth]{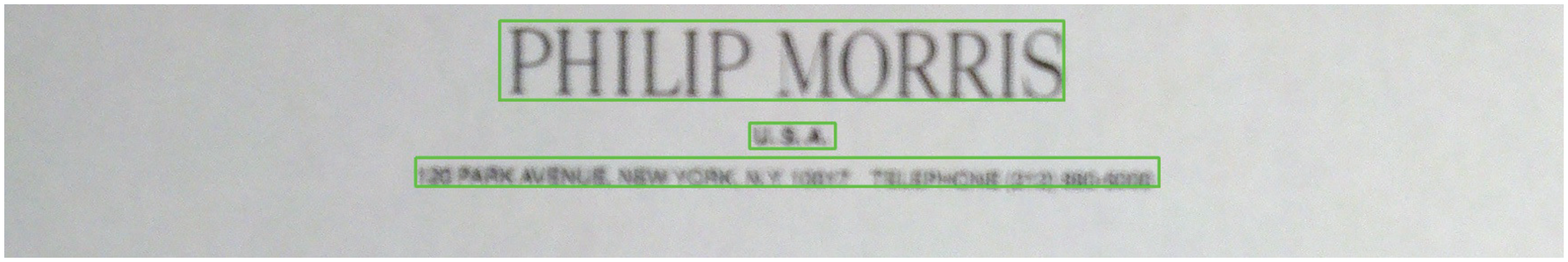}  }
	\caption{Overall quality computed with the weighted pooling and median.} 	
	\label{fig:median}
\end{figure}

An example is shown in Fig.\ref{fig:median}, where the overall quality for these two images was respectively computed with the weighted pooling and median methods. In Fig.\ref{fig:median1}, three text lines were predicted with quality scores of 0.9787, 0.9964, and 0.1087 from top to bottom using our text line quality prediction model. The quality of three text lines in Fig.\ref{fig:median2} corresponded to 0.4845, 0.1345, and 0.1669. The overall document image quality $\hat{q}_{wp}$, obtained with the weighted pooling, was 0.3225 and 0.3736, in agreement with human visual perception according to which both images should have similar quality. But the quality, $\hat{q}_{med}$, computed with the median method, went to both extremes at random, 0.9787 and 0.1669.

Different from the observation of the above example, however, the empirical results on two benchmarks showed that the proposed framework with the median performed better in most cases than with the weighted pooling, as listed in the last two columns of Table \ref{tab:comp:all}. The main reason is that the majority of text lines are close in scale and quality in document images from these two datasets so that the median among text line quality is a better indication of the overall quality of a document image.

\subsection{Comparison with State-of-the-art Methods}

In this experiment, three state-of-the-art DIQA approaches, MetricNR \cite{Nayef2015Metric}, Sharpness \cite{Kumar2012Sharpness}, CG-DIQA \cite{Li2017CGDIQA}, were selected for comparative analysis. They are based on different metrics to compute the quality without prior training. Each method was evaluated on all images over the average accuracy.
Currently, CG-DIQA can achieve the best performance in terms of LCC and SROCC on both benchmarks. We did not compare any CNN-based methods, since a good pretrained model was the key to finetuning on the benchmark and it was publicly unavailable. Moreover, in the finetuning way, an overall evaluation metric was hard to be computed on all images for a fair comparison.

The SmartDoc-QA is a large and complicated dataset, where CG-DIQA can only get the best LCC of 62.50\% and the SROCC of 63.05\%. The proposed framework with weighted pooling (Proposed+WP) achieved better performance than CG-DIQA: 9\% and 11\% increase at LCC and SROCC, as presented in Table \ref{tab:comp:all}.  The LCC and SROCC improved 12\% and 12\% with the Proposed+Median.

The DocImg-QA dataset is relatively simple and small, containing only 175 images. As a result, the metric-based methods can achieve the high LCC and SROCC on this dataset, where the CG-DIQA has got 90.63\% in LCC and 85.65\% in SROCC. The Proposed+WP is slightly superior, nearly 1.5 percent, to the CG-DIQA, as shown in Table \ref{tab:comp:all}. The underlying reasons why the superiority of the proposed method is insignificant on the DocImg-QA dataset are twofold. On one hand, the total number of samples in this dataset is too small and it is hard to further improve performance. On the other hand,
the OCR accuracy as the ground truth quality may not embody the true quality of document images. After all the OCR accuracy that depends on the OCR engine is different from engine to engine.

\subsection{Computational Performance}

The proposed framework consists of three stages, each of which requires different computational cost.
The whole process was running on a server using a single Nvidia GeForce GTX 1080 GPU with an Intel i7-6800k @3.40GHz CPU. In fact, the detection stage was the most time-consuming in the proposed framework. The used detector, CTPN, spent 300 ms on average in finding text lines from a document image. As mentioned previously, it is easy to replace it with another more efficient detector in our framework. The second stage generally cost total 80 ms to predict the quality of all detected text lines in a document image. The ensemble process consumed only around 3 ms on a CPU and thus was negligible.
As a whole, the computational time of the proposed framework required no more than 400ms for each image on the server, which was acceptable in practical applications.

\section{Conclusion}
This paper proposes a text line based DIQA framework, in which the image quality is computed only on detected text lines. To predict text line quality, we synthesized a large-scale text line image dataset and trained a deep network model with it. The proposed framework has strong robustness to heterogeneous document images because 1) text line detection can effectively inhibit the effect of background clutter and noise on the quality assessment, 2) synthetic text line images are diverse enough in attributes, which is greatly beneficial to the CNN model training.

\ifCLASSOPTIONcaptionsoff
\newpage
\fi



\bibliographystyle{IEEEtran}
\bibliography{iqaref}

\begin{thebibliography}{10}
\providecommand{\url}[1]{#1}
\csname url@samestyle\endcsname
\providecommand{\newblock}{\relax}
\providecommand{\bibinfo}[2]{#2}
\providecommand{\BIBentrySTDinterwordspacing}{\spaceskip=0pt\relax}
\providecommand{\BIBentryALTinterwordstretchfactor}{4}
\providecommand{\BIBentryALTinterwordspacing}{\spaceskip=\fontdimen2\font plus
\BIBentryALTinterwordstretchfactor\fontdimen3\font minus
  \fontdimen4\font\relax}
\providecommand{\BIBforeignlanguage}[2]{{%
\expandafter\ifx\csname l@#1\endcsname\relax
\typeout{** WARNING: IEEEtran.bst: No hyphenation pattern has been}%
\typeout{** loaded for the language `#1'. Using the pattern for}%
\typeout{** the default language instead.}%
\else
\language=\csname l@#1\endcsname
\fi
#2}}
\providecommand{\BIBdecl}{\relax}
\BIBdecl

\bibitem{Ye2013Document}
P.~Ye and D.~Doermann, ``Document image quality assessment: A brief survey,''
  in \emph{International Conference on Document Analysis and Recognition},
  2013, pp. 723--727.

\bibitem{Kumar2012Sharpness}
J.~Kumar, F.~Chen, and D.~Doermann, ``Sharpness estimation for document and
  scene images,'' in \emph{International Conference on Pattern Recognition},
  2012, pp. 3292--3295.

\bibitem{Nayef2015Metric}
N.~Nayef, ``Metric-based no-reference quality assessment of heterogeneous
  document images,'' in \emph{SPIE Electronic Imaging}, 2015, pp.
  94\,020L--94\,020L--12.

\bibitem{Li2017CGDIQA}
H.~Li, F.~Zhu, and J.~Qiu, ``{CG-DIQA}: No-reference document image quality
  assessment based on character gradient,'' in \emph{24th International
  Conference on Pattern Recognition (ICPR)}, 2018, pp. 3622--3626.

\bibitem{Kang2014A}
L.~Kang, P.~Ye, Y.~Li, and D.~Doermann, ``A deep learning approach to document
  image quality assessment,'' in \emph{IEEE International Conference on Image
  Processing}, 2014, pp. 2570--2574.

\bibitem{Peng2016Document}
X.~Peng, H.~Cao, and P.~Natarajan, ``Document image quality assessment using
  discriminative sparse representation,'' in \emph{Document Analysis Systems},
  2016, pp. 227--232.

\bibitem{Nayef2015SmartDoc}
N.~Nayef, M.~M. Luqman, S.~Prum, S.~Eskenazi, J.~Chazalon, and J.~M. Ogier,
  ``Smartdoc-qa: A dataset for quality assessment of smartphone captured
  document images - single and multiple distortions,'' in \emph{International
  Conference on Document Analysis and Recognition}, 2015, pp. 1231--1235.

\bibitem{Ye2015Text}
Q.~Ye and D.~Doermann, ``Text detection and recognition in imagery: A survey,''
  \emph{IEEE Transactions on Pattern Analysis and Machine Intelligence},
  vol.~37, no.~7, pp. 1480--1500, 2015.

\bibitem{Tian2016Detecting}
Z.~Tian, W.~Huang, T.~He, P.~He, and Y.~Qiao, ``Detecting text in natural image
  with connectionist text proposal network,'' in \emph{European Conference on
  Computer Vision}, 2016, pp. 56--72.

\bibitem{Buta2015FASText}
M.~Buta, L.~Neumann, and J.~Matas, ``Fastext: Efficient unconstrained scene
  text detector,'' in \emph{IEEE International Conference on Computer Vision},
  2015, pp. 1206--1214.

\bibitem{LiaoSBWL17}
M.~Liao, B.~Shi, X.~Bai, X.~Wang, and W.~Liu, ``Textboxes: A fast text detector
  with a single deep neural network,'' in \emph{AAAI'17}, 2017, pp. 4161--4167.

\bibitem{He:resnet:2016}
K.~He, X.~Zhang, S.~Ren, and J.~Sun, ``Deep residual learning for image
  recognition,'' in \emph{2016 IEEE Conference on Computer Vision and Pattern
  Recognition (CVPR)}, 2016, pp. 770--778.

\bibitem{He:2015:DDR}
------, ``Delving deep into rectifiers: Surpassing human-level performance on
  imagenet classification,'' in \emph{Proceedings of the 2015 IEEE
  International Conference on Computer Vision (ICCV)}, ser. ICCV '15, 2015, pp.
  1026--1034.

\bibitem{Bong2014}
D.~B.~L. Bong and B.~E. Khoo, ``Blind image blur assessment by using valid
  reblur range and histogram shape difference,'' \emph{Signal Processing: Image
  Communication}, vol.~29, no.~6, pp. 699--710, 2014.

\bibitem{Bong2015}
------, ``Objective blur assessment based on contraction errors of local
  contrast maps,'' \emph{Multimedia Tools and Applications}, vol.~74, no.~17,
  pp. 7355--7378, 2015.

\bibitem{Kumar2013A}
J.~Kumar, P.~Ye, and D.~Doermann, \emph{A Dataset for Quality Assessment of
  Camera Captured Document Images}.\hskip 1em plus 0.5em minus 0.4em\relax
  Springer International Publishing, 2013.

\end{thebibliography}
\end{document}